\documentclass[letterpaper, 10 pt, journal]{IEEEtran}  

\IEEEoverridecommandlockouts                              

\usepackage{amsmath} 
\usepackage{amssymb}  
\usepackage{caption}
\usepackage{booktabs}
\usepackage{pifont}
\usepackage{placeins}
\usepackage{afterpage}
\usepackage{graphicx}
\usepackage{subcaption}
\usepackage{array}
\usepackage{booktabs}
\usepackage[table]{xcolor}  
\usepackage{makecell}
\usepackage{multirow}
\usepackage{gensymb}

\usepackage[final]{changes}
\definechangesauthor[color=blue]{SY}

\usepackage{xcolor}
\usepackage{soul}
\sethlcolor{yellow}

\soulregister\cite7
\soulregister\ref7
\soulregister\eqref7
\soulregister\footnote7

\usepackage{xcolor}
\usepackage{tikz}
\usetikzlibrary{calc}
\usepackage{xparse,etoolbox}

\usepackage{xcolor}
\usepackage{soul}
\colorlet{HLyellow}{yellow!60} 
\colorlet{HLcyan}{cyan!20}     
\sethlcolor{HLyellow}

\newcommand{\texthlEdit}[1]{{\sethlcolor{HLcyan}\hl{#1}}}

\newcommand{\mathhl}[2][yellow!60]{%
  \tikz[baseline=(X.base)]{%
    \node[fill=#1, rounded corners=0.5pt, inner xsep=1pt, inner ysep=1pt] (X) {$#2$};%
  }%
}

\newif\ifclean
\cleantrue

\usepackage{tikz}
\usepackage{soul}

\ifclean
  
  \renewcommand{\texthlEdit}[1]{#1}

  \renewcommand{\mathhl}[2][]{\ensuremath{#2}}

\else
  
  \renewcommand{\texthlEdit}[1]{{\sethlcolor{HLcyan}\hl{#1}\sethlcolor{HLyellow}}}

  \renewcommand{\mathhl}[2][yellow!60]{%
    \tikz[baseline=(X.base)]{%
      \node[fill=#1, rounded corners=0.5pt, inner xsep=1pt, inner ysep=1pt] (X) {$#2$};%
    }%
  }

\fi

\usepackage{pdfcomment}
\hypersetup{pdfborder={0 0 0}}

\usepackage{xcolor,pdfcomment}

\title{\LARGE \bf
AFT: Appearance-Based Feature Tracking for Markerless and Training-Free Shape Reconstruction of Soft Robots
}

\ifclean
\author{Shangyuan Yuan, Preston Fairchild, Yu Mei, Xinyu Zhou, and Xiaobo Tan
\thanks{This work was supported in part by the National Science Foundation (CNS 2237577, ECCS 2024649).}
\thanks{All authors are with the Department of Electrical and Computer Engineering, Michigan State University, East Lansing, MI 48824, USA.
        {\tt\small \{yuanshan, fairch42, meiyu1, zhouxi63, xbtan\}@msu.edu}}%
}
\else 
\fi

\usepackage{amsmath}
\begin{document}
\maketitle  

\ifclean
  \thispagestyle{empty}
  \pagestyle{empty}
\else
  \thispagestyle{plain}
  \pagestyle{plain}
\fi

\begin{abstract}

Accurate shape reconstruction is essential for precise control and reliable operation of soft robots.
Compared to sensor-based approaches, vision-based methods offer advantages in cost, simplicity, and ease of deployment. However, existing vision-based methods often rely on complex camera setups, specific backgrounds, or large-scale training datasets, limiting their practicality in real-world scenarios.
In this work, we propose a vision-based, markerless, and training-free framework for soft robot shape reconstruction that directly leverages the robot’s natural surface appearance. These surface features act as implicit visual markers, enabling a hierarchical matching strategy that decouples local partition alignment from global kinematic optimization.
Requiring only an initial 3D reconstruction and kinematic alignment, our method achieves real-time shape tracking across diverse environments while maintaining robustness to occlusions and variations in camera viewpoints.
Experimental validation on a continuum soft robot demonstrates an average tip error of 2.6\% during real-time operation, as well as stable performance in practical closed-loop control tasks. These results highlight the potential of the proposed approach for reliable, low-cost deployment in dynamic real-world settings.

\end{abstract}



\begin{figure*}[t]
    \centering
    \includegraphics[width=\textwidth]{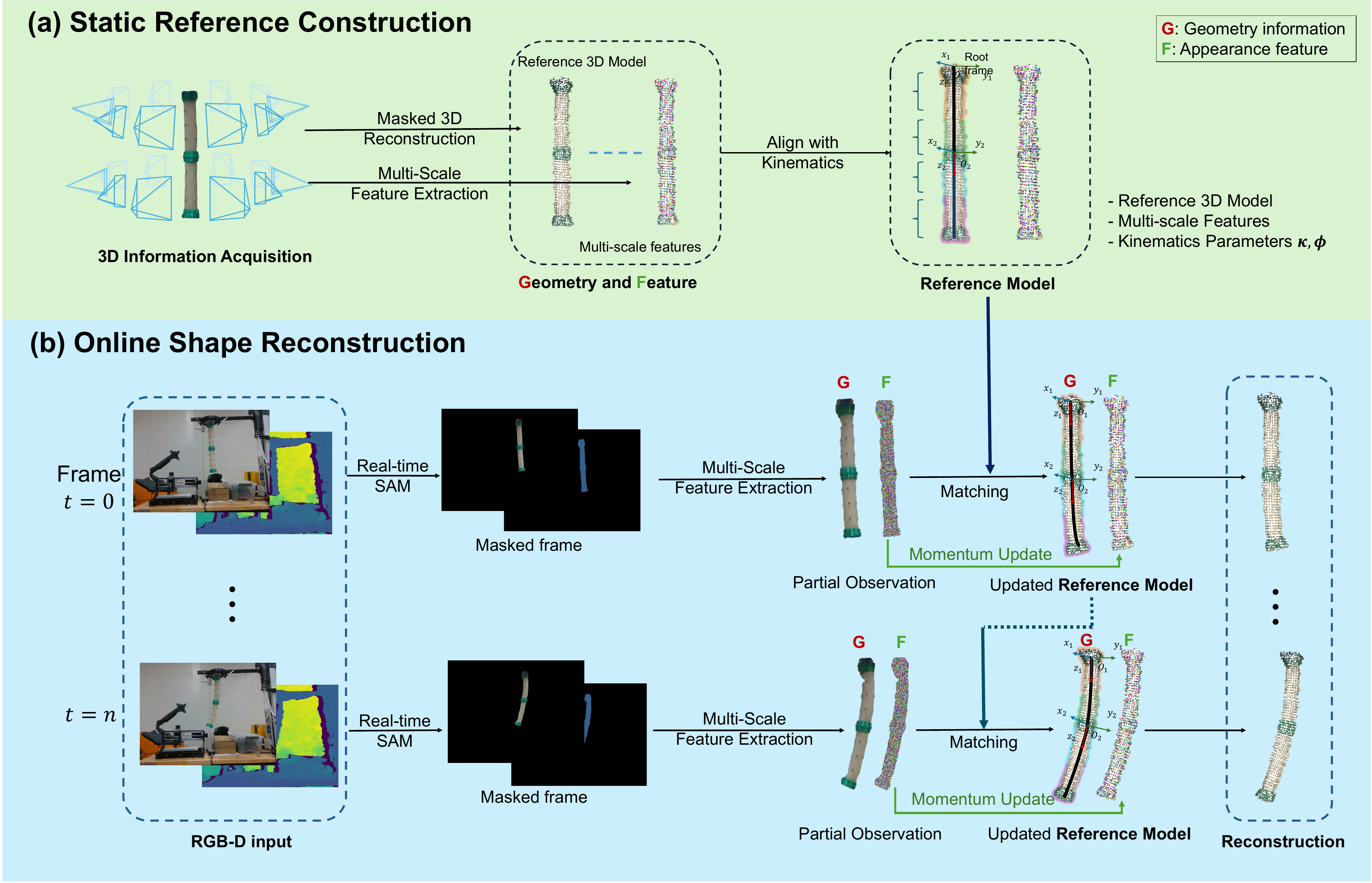}
    \caption{Overview of the proposed framework. The framework is divided into two stages: 
    \textbf{(a) Static Reference Construction}, where a reference model with multi-scale features and backbone kinematics is initialized from multi-view images; and 
    \textbf{(b) Online Shape Reconstruction}, where incoming frames are segmented, matched, and used to update the reference model for continuous 3D shape reconstruction.}
    \label{fig:pipeline}
\end{figure*}

\section{Introduction}

Driven by the increasing demand for physical human–robot interaction, bio-inspired soft robots have gained substantial interest due to their intrinsic compliance and ability to operate safely in contact-rich environments. Their soft structures enable them to perform delicate tasks such as minimally invasive surgery~\cite{runciman2019soft}, fruit harvesting~\cite{elfferich2022soft}, and human assistance~\cite{pan2022soft}. Typically constructed from materials such as silicone, soft robots can undergo high-dimensional and continuous deformations~\cite{della2023model}. While this flexibility benefits applications including navigation and manipulation, it also introduces considerable challenges for reliable shape sensing, highlighting the need for a robust and easily deployable sensing solution.

A widely explored solution is to integrate sensors internally or along the surface of the robot’s body~\cite{mei2024simultaneous}. However, the physical wiring and routing required by these sensing elements can restrict robot deformation and compromise its intrinsic compliance. Moreover, designing and fabricating custom sensing systems can significantly increase fabrication complexity and cost~\cite{li2024unifying, lu2023image}. These challenges have motivated growing interest in vision-based shape sensing approaches, which offer a non-contact alternative that preserves the robot’s compliance.

Vision-based shape sensing methods typically use images as input, often relying on an RGB-D camera or multiple viewpoints to acquire structured 3D information~\cite{wade2022applications}. A crucial preprocessing step is the extraction of robot’s contour, where many approaches assume a strong visual contrast between the robot and the background to enable reliable pixel-level segmentation. After preprocessing, vision-based methods typically reconstruct the robot’s shape by aligning a kinematic model with visual observations. Their differences mainly lie in two aspects: the kinematic modeling of continuum soft robots and the optimization paradigm.

Regarding the kinematic modeling, different priors have been employed, such as geometric strain bases~\cite{albeladi2021vision}, clothoid curves~\cite{mei2025learning}, and piecewise constant-curvature segments~\cite{schindler2024image}. In contrast, some learning-based methods bypass explicit kinematic modeling altogether and directly predict the 3D positions of $N$ points along the robot backbone~\cite{zheng2024vision}. 
With respect to optimization, a common technique is Iterative Closest Point (ICP)~\cite{hoffmann2024iterative}. Differentiable rendering has also been explored to improve efficiency~\cite{li2024unifying}, but the process remains time-consuming. In contrast, learning-based methods use neural networks to directly infer the robot’s shape from images~\cite{reiter2011learning,lal2021scoopnet}, offering fast and accurate performance. However, they require large and rich labeled datasets, which is particularly challenging given the complex and continuous deformations of soft robots.

Despite recent advances, rapid deployment of existing methods remains a significant obstacle. These methods often impose strict constraints on camera placement~\cite{li2024unifying}, rely on clear contours or clean backgrounds, or require large labeled datasets. Such constraints limit their robustness to complex conditions such as occlusions~\cite{rong2024vision}, environmental variations, and viewpoint changes~\cite{zheng2024vision}, which are commonly encountered in real-world scenarios such as agricultural environments~\cite{magistri2022contrastive}. These challenges call for a deployment-friendly approach capable of operating reliably under diverse real-world conditions. 

An intuitive strategy to improve visual observability and tracking robustness is to physically attach identifiable markers onto the robot’s surface. However, maintaining the visibility of traditional fiducial markers requires multi-camera tracking systems~\cite{zhang2019calibration,bern2020soft,huang2021kinematic}, as missing even a subset of markers can cause ambiguity in tracking due to their lack of unique identity. Alternative approaches, such as using ArUco markers~\cite{kara2023towards,albeladi2022hybrid}, are prone to failure under large deformations, as the marker patterns are easily distorted beyond recognition. 
A more ideal strategy would be to densely distribute distinct, easily recognizable color patches across the robot’s surface, enabling robust correspondence even under occlusions and deformations. Such a design, however, would significantly complicate fabrication, maintenance, and deployment.

A more interesting idea is to leverage the inherent surface textures of the robot as implicit visual markers. On the contrary to traditional markers, ours are both \emph{unique} and \emph{abundant}. Even if a large portion of them is occluded, the remaining is still sufficient to support accurate reconstruction of the robot’s shape. Building on this insight, we propose Appearance-based Feature Tracking~(AFT), a vision-based, markerless, training-free (i.e., no task-specific training or fine-tuning required) framework for soft robots' shape reconstruction.

Our main contribution is a vision-based, markerless, and training-free framework for real-time soft robot shape reconstruction. Leveraging the robot’s natural surface appearance as implicit visual markers, the method enables robust feature matching. Then shape reconstruction is achieved through a hierarchical reconstruction strategy that decouples local partition matching from global kinematic optimization. The framework requires only an initial 3D reconstruction and kinematic alignment, and operates with a single uncalibrated RGB-D camera under minimal placement constraints. It is robust to challenging conditions such as varying backgrounds, lighting, viewpoints, and partial occlusions, enabling rapid deployment in unstructured environments. We also validate the method on a soft continuum robot in a closed-loop feedback control task, demonstrating stable and responsive deformation regulation.




An overview of the framework is shown in Fig.~\ref{fig:pipeline}. The remainder of this paper is organized as follows. Section~\ref{sec:static_shape_reconstruction} introduces the preparation step, where a static reference model is constructed using multi-view observations and kinematic alignment. Section~\ref{sec:realtime_feature_matching} then describes the online stage, where incoming RGB-D frames are processed in real time to update the reference model for continuous shape tracking. Section~\ref{sec:experiments} presents experimental validation and analysis, and Section~\ref{sec:conclusion} concludes the paper and discusses future directions.

\section{Static Reference Construction}
\label{sec:static_shape_reconstruction}

This section introduces the construction of a static Reference Model, as shown in Fig.~\ref{fig:pipeline}~(a). Conceptually, it serves a role similar to placing physical markers on the robot—providing trackable reference points—but instead relies on distinctive surface appearance features.

\subsection{Masked 3D Reconstruction}
To construct the geometric component of the Reference Model, we perform dense 3D reconstruction of the undeformed robot using the COLMAP~\cite{schonberger2016structure}. Images are captured from multiple viewpoints, producing a dense point cloud, intrinsic and extrinsic camera parameters, and per-frame depth maps. 
To isolate the robot from the background, we perform per-frame segmentation using the Segment Anything Model (SAM)~\cite{ravi2024sam}. We provide a bounding-box prompt in an initial reference frame and additional point prompts to later frames to refine segmentation accuracy. Then the segmented surface points are downsampled using farthest point sampling (FPS)~\cite{qi2017pointnet++}.

\subsection{Multi-Scale Feature Extraction}
This step aims to enrich each 3D point in the Reference Model with a distinctive descriptor. Each 3D surface point is associated with its corresponding image pixels via back-projection. We then extract multi-scale features using a ResNet-50 backbone~\cite{he2016deep} pretrained on ImageNet~\cite{deng2009imagenet}. As shallower layers capture local textures while deeper layers encode semantic context~\cite{lin2017feature}, we extract feature maps from multiple intermediate layers with receptive fields of $7\times7$, $35\times35$, and $99\times99$, as illustrated in Fig.~\ref{fig:feature-receptive-field}.
\texthlEdit{All feature maps are resized to the same spatial resolution so that the projected pixel coordinate is aligned across scales }\mathhl[HLcyan]{s \in \mathcal{S}}\texthlEdit{. To aggregate the multi-view observations, we select the descriptor with the highest average similarity to the rest:}{
\begin{equation}
{f}^{(s)}_i = \arg\max_{f \in \{f^{(s)}_{i,j}\}_{j=1}^{V_i}} \frac{1}{V_i} \sum_{k=1}^{V_i} \cos\left(f, f^{(s)}_{i,k}\right),
\end{equation}
where \(\{f^{(s)}_{i,j}\}_{j=1}^{V_i}\) denotes the set of feature vectors corresponding to point \(p_i\) observed in \(V_i\) views at scale \(s\), and \(\cos(\cdot, \cdot)\) denotes the cosine similarity, a commonly used metric for measuring similarity between feature vectors~\cite{he2020momentum}, defined as:
\begin{equation}
\cos(\mathbf{u}, \mathbf{v}) = \frac{\mathbf{u} \cdot \mathbf{v}}{\lVert \mathbf{u} \rVert \, \lVert \mathbf{v} \rVert},
\end{equation}
where \( \|\cdot\| \) denotes the Euclidean norm. \texthlEdit{The final feature descriptor }\mathhl[HLcyan]{\hat{f}_i}\texthlEdit{ for each point $p_i$ is obtained by concatenating its scale-specific features}, \(\hat{f}_i = \left[ \, {f}_i^{(s)} \,\right]_{s \in \mathcal{S}}\).

\begin{figure}[htbp]
    \centering
    \includegraphics[width=0.7\linewidth]{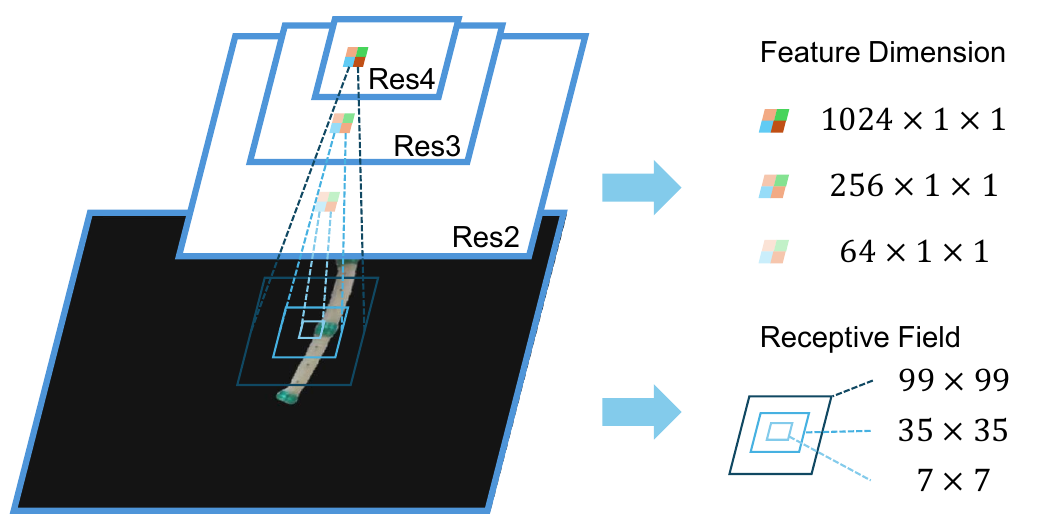}
    \caption{Illustration of multi-scale feature extraction. Features are extracted from Res2, Res3, and Res4 stages of a ResNet-50, providing increasing receptive fields and feature dimensions for hierarchical representation.}
    \label{fig:feature-receptive-field}
\end{figure}

\subsection{Kinematics Alignment}

After obtaining the geometry and associated features, we establish a mapping between surface points and the robot’s kinematic state. We assume that the deformation of the robot backbone can be described by a predefined forward kinematics (FK) model. Specifically, given a configuration vector \(\boldsymbol{\xi}\), the model provides the position \(t(\boldsymbol{\xi}, \sigma)\) and orientation \(R(\boldsymbol{\xi}, \sigma)\) of a center-line point at coordinate \(\sigma\). The configuration vector \(\boldsymbol{\xi}\) encodes actuation or deformation parameters, while \(\sigma\) denotes a structural coordinate such as arc length, depending on the FK model.
With the above definitions, we introduce two complementary components:

\subsubsection{Surface Point Kinematics}
Given the backbone configuration \(\boldsymbol{\xi}\), the 3D position of each surface point \(p_i\) is computed as:
\begin{equation}
p_i(\boldsymbol{\xi}) = t(\boldsymbol{\xi}, \sigma_i) + R(\boldsymbol{\xi}, \sigma_i) \cdot \left( p_i^0 - t(\boldsymbol{\xi}^{(0)}, \sigma_i) \right),
\end{equation}
where \(\sigma_i\) denotes the structural coordinate of the backbone point closest to \(p_i\) in the undeformed configuration, and \(p_i^0\) is the corresponding surface point position in that state. The term \(\boldsymbol{\xi}^{(0)}\) refers to the backbone configuration in the undeformed state.

\subsubsection{Backbone-Driven Inverse Kinematics}

Given a set of target positions \(\{p_{\text{target}}^{(j)}\}_{j=1}^C\) defined at known structural coordinates \(\{\sigma_j\}_{j=1}^C\) along the backbone, we estimate the configuration parameters \(\boldsymbol{\xi}\) by minimizing the cumulative deviation between the predicted and target positions:
\begin{equation}
\boldsymbol{\xi}^* = \arg\min_{\boldsymbol{\xi} \in \mathcal{W}} \sum_{j=1}^C \left\| p_{\text{target}}^{(j)} - t(\boldsymbol{\xi}, \sigma_j) \right\|^2,
\label{eq:ik_backbone}
\end{equation}
where \(\{\sigma_j\}_{j=1}^C\) are the structural coordinates at which the target positions are defined, and \(t(\boldsymbol{\xi}, \sigma_j)\) denotes the predicted position at coordinate \(\sigma_j\) based on the forward kinematics model. \(\mathcal{W}\) denotes the feasible domain of the configuration vector \(\boldsymbol{\xi}\).

\subsection{Initial Reference Model}

The initial reference model, as illustrated in Fig.~\ref{fig:pipeline}~(a), provides a geometric and kinematic framework that enables forward shape prediction and state estimation via inverse kinematics. 
During online reconstruction, it serves as the anchor for associating live observations and tracking the robot’s shape efficiently.

\section{Online Shape Reconstruction}
\label{sec:realtime_feature_matching}

This section aims to estimate the robot’s deformation from an incoming RGB-D video stream in real time. As shown in Fig.~\ref{fig:pipeline}~(b), we update the reference model after \texthlEdit{each frame and match the subsequent frame to this updated model}.

\subsection{Data Processing}

We first segment the robot in the RGB image using a lightweight open-source implementation of SAM~\cite{gy920_segment_anything_2024}. The resulting binary mask is projected to the depth map to extract 3D points, followed by density-based filtering to remove noise and segmentation artifacts. Multi-scale visual features are extracted from the masked RGB image 
as in Section~\ref{sec:static_shape_reconstruction}. This produces a set of feature-embedded surface points, which serve as input for the matching process described next.

\subsection{Real-Time Feature Matching}

We consider both feature similarity and spatial proximity. Feature similarity is computed via cosine similarity across multi-scale features. Spatial proximity accounts for geometric consistency based on 3D positions.
We define the overall matching score between a reference point $p_i \in \mathcal{P}_{\text{obs}}$ and an observed surface point \( q_j \in \mathcal{Q} \) as:
\begin{equation}
    S_{\text{match}}(p_i, q_j) = \cos\left( \hat{f}_{p_i}, \hat{f}_{q_j} \right) \cdot \exp\left( -\frac{D(p_i, q_j)^2}{\sigma^2} \right),
\end{equation}
where $\mathcal{P}_{\text{obs}} = \{p_i\}_{i=1}^{n_{\text{obs}}}$ and ${\mathcal{Q} = \{q_j\}_{j=1}^{n_{\text{live}}}}$ denote the sets of visible reference points and current observed surface points, respectively. ${\mathcal{P}_{\text{obs}}}$ is obtained by projecting the 3D points onto the image plane with the camera intrinsics, while retaining only the depth-closest point for each pixel. $\sigma$ is the bandwidth parameter of the Gaussian kernel, controlling the decay rate with respect to the distance.
The score matrix \(S_{\text{match}} \in \mathbb{R}^{n_{\text{live}} \times n_{\text{obs}}}\) 
captures the pairwise matching likelihoods between these two sets.
The multi-scale cosine similarity is defined as:
\begin{equation}
    \cos(\hat{f}_{p_i}, \hat{f}_{q_j}) = \frac{1}{S} \sum_{s=1}^{S} \cos\left(f_{p_i}^{(s)}, f_{q_j}^{(s)}\right),
\end{equation}
and the spatial distance is computed as:
\begin{equation}
    D(p_i, q_j) = \left\| \mathbf{X}_{p_i} - \mathbf{X}_{q_j} \right\|,
\end{equation}
where \( \mathbf{X}_{p_i} \) is the stored 3D position of \( p_i \) and \( \mathbf{X}_{q_j} \) is that of \( q_j \), derived from depth sensing.

To determine optimal correspondences, we solve a maximum-weight bipartite matching problem that maximizes total matching score under a one-to-one constraint:
\begin{equation}
    \mathcal{M}^* = \arg\max_{\mathcal{M} \in \mathbb{M}} \sum_{(p_i, q_j) \in \mathcal{M}} S_{\text{match}}(p_i, q_j),
\end{equation}
where \(\mathbb{M}\) denotes the set of all feasible matchings with no duplicate assignments, 
and we resolve this assignment problem by applying the Hungarian algorithm~\cite{kuhn1955hungarian}.

To account for appearance changes over time, we update the feature representation of each matched reference point. For each matched pair \((p_i, q_j) \in \mathcal{M}^*\), the update rule is:
\begin{equation}
    \hat{f}_{p_i} \leftarrow (1 - \alpha) \cdot \hat{f}_{p_i} + \alpha \cdot \hat{f}_{q_j},
\end{equation}
where \( \alpha \in [0, 1] \) controls the momentum of the update.

\subsection{Shape Reconstruction}
Traditional marker-based methods reconstruct shape by minimizing the error between observed and predicted marker positions. However, such optimization remains computationally expensive even with a limited number of markers. Therefore, we propose a \textit{hierarchical shape reconstruction strategy} that decouples local partition-based matching from global kinematic parameter estimation. Specifically, we approximate the deformable robot shape as a chain of partitions, as illustrated in the “Reference Model” of Fig.\ref{fig:pipeline}(a) with partitions differentiated by colors. We assume that each partition remains rigid only between two consecutive frames as a short-term approximation. Each partition is estimated independently and subsequently integrated through a global optimization.

\begin{figure*}[t]
    \centering
    \includegraphics[width=\textwidth]{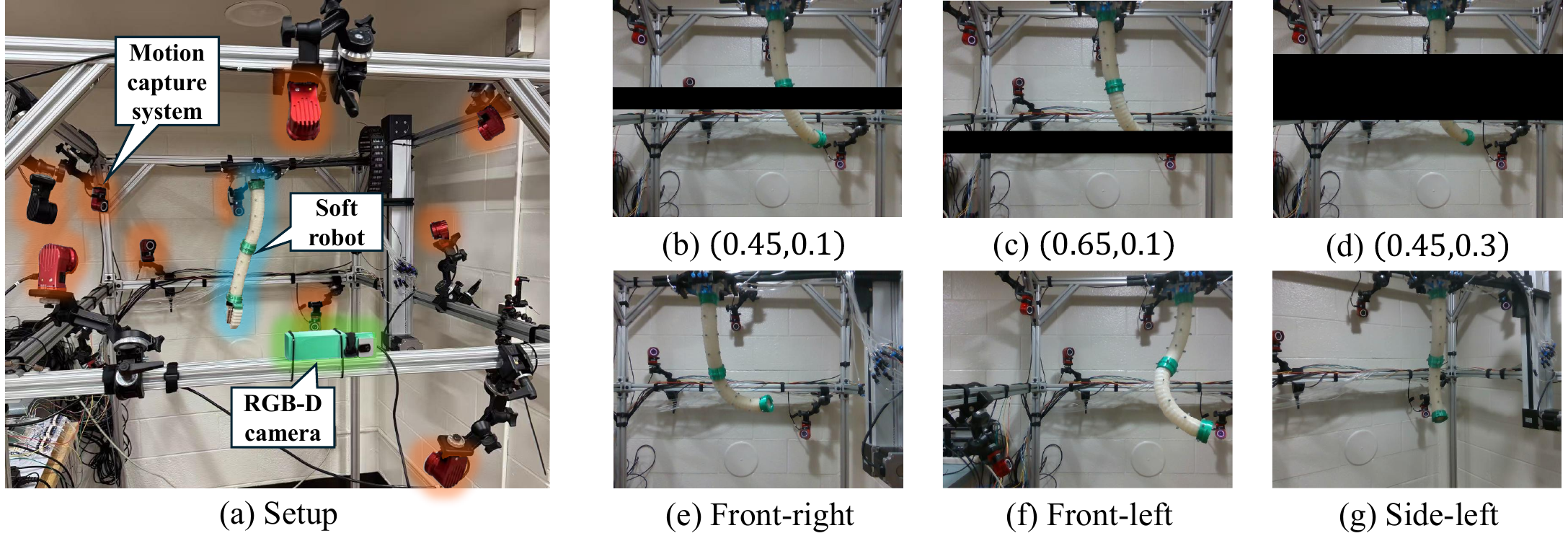}
    \caption{
    Illustration of experimental setups used to evaluate robustness under occlusion and varying viewpoints.
    (a) shows the overall experimental setup, including an RGB-D camera, a soft continuum robot, and a motion capture system.
    (b)--(d) depict occlusion settings with horizontal bars digitally introduced at different (position, width) values, where the first number denotes the normalized height along the image and the second denotes the relative width (0 indicates no occlusion).
    (e)--(g) show images captured from different viewpoints: front-right, front-left, and side-left.
    }
    \label{fig:combined_results}
\end{figure*}

\subsubsection{Local Partition-Based Transformation Estimation}

Each partition, denoted \( P^{(j)} \ (j = 1, \dots, K) \), represents a local region over which a transformation (translation and rotation) is estimated to describe its current position and orientation. We consider the subset of matched pairs \( (p_m, q_n) \in \mathcal{M}_j \), where \( \mathcal{M}_j \subseteq \mathcal{M}^* \) collects the assignments whose reference points \( p_m \) belong to partition \( P^{(j)} \).  
The transformation \( \mathbf{T}_j^* = [\mathbf{R}_j \mid \mathbf{t}_j] \) of partition \( j \) is estimated by:
\begin{equation}
    \mathbf{T}_j^* = \arg\min_{\mathbf{T}_j} \sum_{(p_m, q_n) \in \mathcal{M}_j} \left\| \mathbf{R}_j \mathbf{X}_{p_m} + \mathbf{t}_j - \mathbf{X}_{q_n} \right\|^2,
\end{equation}
where \( \mathbf{R}_j \in \mathrm{SO}(3) \) and \( \mathbf{t}_j \in \mathbb{R}^3 \) denote the rotation and translation components of the transformation, respectively; \( \mathbf{X}_{p_m}, \mathbf{X}_{q_n} \in \mathbb{R}^3 \) are the 3D coordinates of \( p_m \) and \( q_n \), respectively.

\subsubsection{Global Backbone Parameter Optimization}

The optimization minimizes the deviation between the transformed base points obtained from local matching and those predicted by the global forward kinematics:
\begin{equation}
\boldsymbol{\xi}^* = \arg\min_{\boldsymbol{\xi} \in \mathcal{W}} \sum_{j=1}^{K} \left\| \mathbf{T}_j(\mathbf{X}_{\sigma_j}) - t(\boldsymbol{\xi}, \sigma_j) \right\|^2,
\end{equation}
where \(\mathbf{T}_j(\mathbf{X}_{\sigma_j})\) denotes the position of the base point \(\mathbf{X}_{\sigma_j}\) after applying the estimated local transformation for partition \(P^{(j)}\), and \(t(\boldsymbol{\xi}, \sigma_j)\) is the predicted position at coordinate \(\sigma_j\) given the current configuration.  
The refined backbone parameters \(\boldsymbol{\xi}^*\) are then used to reconstruct the global robot shape.

Collectively, these modules form a real-time shape reconstruction pipeline that extends naturally from per-frame processing to sequential tracking, as illustrated in Fig.~\ref{fig:pipeline}~(b).

\section{Experiments and results}
\label{sec:experiments}

\begin{table*}[t]
\centering
\caption{Comparison of existing soft robot shape reconstruction methods. Our method is contour/background agnostic, requires no fixed camera setup, markers, or training data, while achieving competitive accuracy.}
\label{tab:comparison}

\renewcommand{\arraystretch}{1.2}
\begin{tabular}{>{\centering\arraybackslash}m{2.4cm} 
                >{\centering\arraybackslash}m{1.4cm} 
                >{\centering\arraybackslash}m{2.79cm} 
                >{\centering\arraybackslash}m{1.8cm} 
                >{\centering\arraybackslash}m{1.2cm} 
                >{\centering\arraybackslash}m{1.8cm} 
                >{\centering\arraybackslash}m{1.7cm} 
                >{\centering\arraybackslash}m{1.3cm}}
\toprule
\makecell[c]{\textbf{Method}} & 
\makecell[c]{\textbf{Camera} \\ \textbf{Setup}} &
\makecell[c]{\textbf{Contour / Background} \\ \textbf{Agnostic}} &
\makecell[c]{\textbf{Fixed} \\ \textbf{Camera Setup}} &
\makecell[c]{\textbf{Markers} \\ \textbf{Required}} &
\makecell[c]{\textbf{Training Data} \\ \textbf{Required}} &
\makecell[c]{\textbf{Robot Length} \\ \textbf{(mm)}} &
\makecell[c]{\textbf{Tip Error} \\ \textbf{(\%)}} \\
\midrule

Hehui et al.~\cite{zheng2024vision}      & 2$\times$RGB& No  & Yes & No  & Yes & 335 & $3.6 \pm 5.0$\textsuperscript{$\dagger$} \\
Yu et al.~\cite{rong2024vision}         & 3$\times$RGB & No  & Yes & No  & Yes & 175 & 1.3 \\
Camarillo et al.~\cite{camarillo2008vision}  & 3$\times$RGB & No  & Yes & No  & No & 160 & 4.8 \\
Vandini et al.~\cite{vandini2017unified}    & 1$\times$RGB  & No  & Yes & No  & No  & 260 & 2.8 \\
Pedari et al.~\cite{pedari2019spatial}   & 2$\times$RGB     & No  & Yes  & Yes  & No  & {--} & 4.5 \\
Ali et al.~\cite{albeladi2021vision}        & RGB-D   & No  & Yes & No  & No & 287 & $4.5 \pm 3.1$ \\
Jingpei et al.~\cite{lu2023image}    & RGB-D   & No  & Yes & No  & No & 200 & $3.5 \pm 1.1$\textsuperscript{$\ddagger$} \\
\rowcolor{cyan!10}
\textbf{Ours}     & RGB-D   & \textbf{Yes} & \textbf{No} & \textbf{No} & \textbf{No} & 400 & \textbf{$2.6 \pm 1.3$} \\
\bottomrule
\end{tabular}

\vspace{1mm}
\begin{minipage}[t]{0.97\textwidth}
\raggedright
\small
\textit{Note:} “Contour / Background Agnostic” indicates that the method does not require precise contour segmentation or a clean background. \textsuperscript{$\dagger$} The lower error ($0.3 \pm 0.2$\%) is achieved with additional surface feature tracking; the reported value reflects the standard version. \textsuperscript{$\ddagger$} As tip error is not reported in~\cite{lu2023image}, the overall average error is used instead.
\end{minipage}
\end{table*}

\subsection{Implementation Details}

The experimental setup is shown in Fig.~\ref{fig:combined_results}~(a). 
The soft robot used in our experiments is a pneumatically actuated two-segment manipulator fabricated from Smooth-On Dragon Skin\textsuperscript{TM} 30 silicone. 
Each segment measures 20~cm in length and 4~cm in diameter, resulting in a total length of 40~cm. 
Each segment contains three internal air chambers spaced 120\degree{} apart, with actuation pressures up to 100~kPa.
To represent the robot's configuration, we adopt the Piecewise Constant Curvature (PCC) model~\cite{webster2010design}, where each soft segment is modeled as a circular arc that is parameterized by curvature \(\kappa_i\), bending direction \(\phi_i\), and arc length \(l_i\). These parameters form the segment-wise configuration vector \(\boldsymbol{\xi}_i = (\kappa_i, \phi_i, l_i)\), and the complete robot configuration is denoted as \(\boldsymbol{\xi} = [\boldsymbol{\xi}_1, \boldsymbol{\xi}_2]\).
The arc lengths \(l_i\) are estimated from the chamber pressures \(P_i = [P_{i,1}, P_{i,2}, P_{i,3}]^\top\) using a linear regression model:
\begin{equation}
l_i = k_i^\top P_i + l_{0,i},
\label{eq:pressure_length_mapping}
\end{equation}
where \(k_i\) is the coefficient vector and \(l_{0,i}\) is the nominal arc length at zero pressure. We collected additional 50 pairs of pressure–length samples to fit these parameters. Regarding radial expansion, prior work~\cite{habibian2022evaluation} has shown that \texthlEdit{when the internal pressure is no higher than 100~kPa}, the expansion remains below 2\%. We neglect this effect in our experiments.

\textit{Static Reference Model.} 
We capture 31 high-resolution images ($718 \times 1278$) using a mobile phone to construct the undeformed point cloud (1,163,125 points), which is then downsampled to 1,992 points. 
Feature maps from the 2\textsuperscript{nd}, 5\textsuperscript{th}, and 8\textsuperscript{th} layers of a pre-trained ResNet-50 are used, with dimensions of 64, 256, and 1024, and receptive fields of $7 \times 7$, $35 \times 35$, and $99 \times 99$ pixels. 

\textit{Online Shape Reconstruction.}
We use an Intel RealSense D435 RGB-D camera with an image resolution of $480 \times 640$. 
Images are resized to $192 \times 256$ before normalization. 
For local transformation estimation, the reference model is partitioned into \(K = 4\) segments of approximately equal arc length. 
During feature update, we apply a Gaussian kernel with $\sigma=1$ and an update rate of $\alpha=0.1~\mathrm{s}^{-1}$.
All online reconstruction experiments are performed on a workstation running Ubuntu 22.04 with an Intel Core i9-12900 CPU and an NVIDIA RTX A4000 GPU (16~GB), using CUDA 12.6.

For the main experiment reported in Table~\ref{tab:comparison}, we collected 50 actuation sets by randomly assigning pressures to the six chambers, with each chamber having a 30\% chance of being set to zero.
To obtain accurate ground-truth, we employ an optical motion capture system (OptiTrack). A total of 36 reflective markers are attached to the robot surface, and their positions are grouped to compute 9 backbone reference points.

\subsection{Benchmarks}
We compare our method against several related works on shape reconstruction for continuum soft robots, as summarized in Table~\ref{tab:comparison}. These approaches often rely on markers or specialized camera setups and are evaluated on different platforms. Due to the lack of publicly available datasets or source code, we restrict the comparison to the reported relative tip error—commonly used as the primary evaluation metric, since the tip typically exhibits the largest deformation in continuum robots. Moreover, as the tip is frequently used to attach end-effectors for tasks such as grasping, this metric is also of practical significance. To ensure fairness, all reported errors are normalized by the corresponding robot length. As runtime speed is typically influenced by hardware differences, we do not include execution frequency.

\subsection{Results and Analysis}

Our method achieves a relative tip error of $2.6 \pm 1.3\%$ at an update rate of 2.5~Hz using a single RGB-D camera. This performance is attained without reliance on fixed camera setups, markers, training data, or assumptions about the background. 

In our setup, The green components and the subtle surface textures from fiber reinforcement provide sufficient texture variation.
\texthlEdit{For a fair comparison, we do not include surface bands, even though they can substantially reduce errors}~\cite{zheng2024vision}. Due to the near-uniform surface texture, rotation about the cylindrical axis remains ambiguous.

Unlike prior works that assume a clean background or a known mask (e.g., via colored backdrops~\cite{zheng2024vision}), we obtain the robot mask with SAM. Although SAM runs at $\sim$5~Hz and thus bounds the update rate, it provides robust segmentation. Importantly, SAM and ResNet are used only as off-the-shelf preprocessing; our contributions lie in the reference model, the multi-scale feature extraction that yields identifiable markers, and the hierarchical reconstruction strategy enabling efficient inference.

\subsection{Ablation Studies}

To further evaluate the contribution of each module, we conduct ablation experiments on the same 50 test sequences. For efficiency, the sequences are downsampled to 1 Hz (about 15 frames per sequence).

The results are summarized in Table~\ref{tab:ablation}. The full framework achieves a tip error of 2.60\%, which is close to the result obtained at the original 2.5 Hz (runtime $\approx$ 0.4 s/frame). Notably, the shape-wise error is higher than the tip error. We attribute this to the robot’s appearance as shown in Fig.~\ref{fig:combined_results}: the tip is a green cylinder with stronger texture distinctiveness than the body, which allows for more accurate matching. This is also consistent with the “w/o multi-scale feature extraction” configuration (geometry-only matching), where the overall shape-wise error is slightly lower but the tip error becomes higher.
Removing the reference feature map update results in only a marginal performance drop, suggesting that the update mechanism has limited effect in short sequences.
Finally, removing the hierarchical reconstruction strategy (i.e., using ICP and traditional optimization only) leads to a significant performance degradation: both tip and shape-wise errors increase considerably, and the runtime increases drastically to 9.5 s/frame ($\approx$ 0.1 Hz).

\begin{table}[htbp]
\scriptsize
\centering
\caption{Ablation study on different components of our framework. 
Errors are reported as mean relative tip and shape-wise error.}
\label{tab:ablation}
\resizebox{\columnwidth}{!}{%
\begin{tabular}{lccc}
    \toprule
    \textbf{Configuration} & \textbf{Tip (\%)} & \textbf{Shape (\%)} & \textbf{Runtime (s/frame)} \\
    \midrule
    Full framework (all modules)        & 2.60 & 2.85 & 0.4 \\
    w/o multi-scale feature extraction  & 2.84 & 2.73 & 0.4 \\
    w/o reference feature map update    & 2.63 & 2.92 & 0.4 \\
    w/o hierarchical reconstruction    & 2.99  & 2.86 & 9.5 \\
    \bottomrule
\end{tabular}
}
\end{table}

\begin{figure}[htbp]
    \centering
    \begin{subfigure}[b]{0.48\columnwidth}
        \centering
        \includegraphics[width=\linewidth]{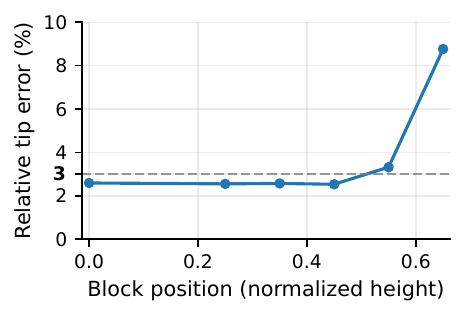}
        \caption{Block positions.}
        \label{fig:occlusion_positions}
    \end{subfigure}
    \hfill
    \begin{subfigure}[b]{0.48\columnwidth}
        \centering
        \includegraphics[width=\linewidth]{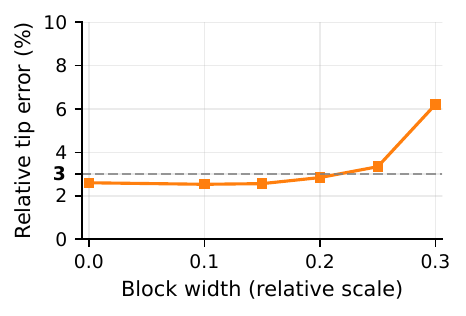}
        \caption{Block widths.}
        \label{fig:occlusion_widths}
    \end{subfigure}
    \caption{Performance under different occlusion settings. 
    (a) Relative tip error under varying block positions. 
    (b) Relative tip error under varying block widths.}
    \label{fig:occlusion_results}
\end{figure}

\begin{figure}[htbp]
    \centering
    \includegraphics[width=0.4\columnwidth]{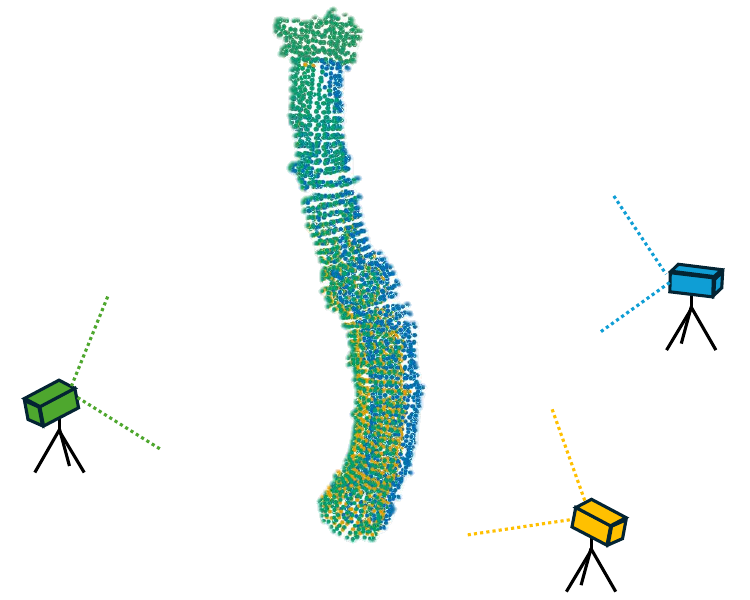}
    \caption{Experimental demonstration of viewpoint robustness. 
    The same target shape is reconstructed from three viewpoints (blue, orange, green) and visualized in a common frame (via calibrated extrinsics), showing close overlap. }
    \label{fig:viewpoint_invariance}
\end{figure}

\begin{figure*}[htbp]
    \centering
    \includegraphics[width=\textwidth]{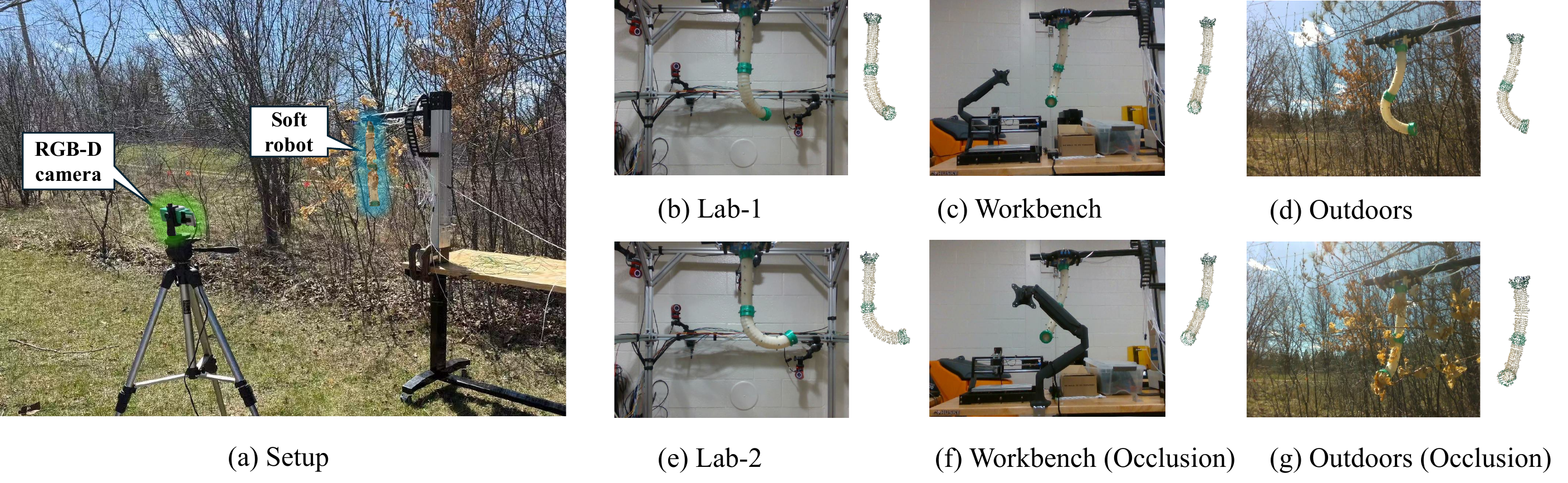}
    \caption{Experimental setup and qualitative demonstration of our method across multiple real-world scenarios. (a) shows the outdoor experimental setup, consisting of an RGB-D camera and a soft continuum robot mounted on a workbench. (b)--(g) present representative results from indoor and outdoor environments, both with and without occlusions. While only RGB images are shown for visualization (left), both color and depth data are used for 3D shape reconstruction (right). Partial occlusions are introduced in (f) and (g) to simulate realistic deployment challenges in cluttered environments.
    }
    \label{fig:outdoor_demo}
\end{figure*}

\subsection{Occlusion}
We adopt and extend the occlusion setup proposed in~\cite{zheng2024vision}, where occlusions were introduced only near the middle and tip of the robot. Here, we digitally insert horizontal blocks into both RGB and depth channels of 50 test sequences 
(Fig.~\ref{fig:combined_results}(b)--(d)). Two experiments are performed: (i) varying the block position along normalized image height 
($0=$ none, larger values closer to the tip), and (ii) varying the block width as a ratio of image width 
(centered at the robot, $0=$ none, $1.0=$ full width). 
As shown in Fig.~\ref{fig:occlusion_results}, our method remains robust under moderate occlusions 
(position $<0.55$, width $<0.25$).

\subsection{Variable Camera Viewpoints}

We test our method at three additional viewpoints (front-right, front-left, and side-left) under identical conditions (Fig.~\ref{fig:combined_results}(e)--(g)). For each viewpoint, three sequences are recorded, for nine in total. A qualitative visualization of this setup is shown in Fig.~\ref{fig:viewpoint_invariance}.
As shown in Table~\ref{tab:camera_viewpoints}, the relative tip error averages $2.66 \pm 0.57\%$, close to the $2.60\%$ under the default 50-sequence setting, demonstrating robustness to moderate changes in camera placement.

\begin{table}[htbp]
\centering
\caption{Relative tip error under different viewpoints}
\label{tab:camera_viewpoints}
\setlength{\tabcolsep}{4pt}
\begin{tabular}{l c c c c}
    \toprule
    \textbf{Metric} & \textbf{Front-right} & \textbf{Front-left} & \textbf{Side-left} & \textbf{Overall Avg.} \\
    \midrule
    Tip Error (\%) & $2.8 \pm 0.8$ & $2.6 \pm 0.5$ & $2.6 \pm 0.4$ & $\mathbf{2.7 \pm 0.6}$ \\
    \bottomrule
\end{tabular}
\end{table}

\subsection{Closed-loop Control}

To evaluate the effectiveness of our method in real-time control, we conduct two closed-loop control tasks: shape control and tip position control. In the shape control task, the robot tracks a target configuration specified by two-segment parameters \((\boldsymbol{\kappa}, \boldsymbol{\phi})\), with performance evaluated by the average error across eight centerline points. In the tip control task, a 3D target position is specified, and the corresponding bending parameters are computed via inverse kinematics (Section~\ref{sec:static_shape_reconstruction}), with performance measured by the 3D position error of the tip. Each task is evaluated over five test sequences, with errors averaged across the final five frames after the robot reaches steady state.
The controller used in our experiments is shown in Fig.~\ref{fig:controller}.
It computes the air pressure vector \( P \in \mathbb{R}^6 \) based on the estimated curvature and bending direction.
The control logic enforces spatial sparsity by modulating pressure via an angular weighting function.

\begin{figure}[t]
    \centering
    \includegraphics[width=\linewidth]{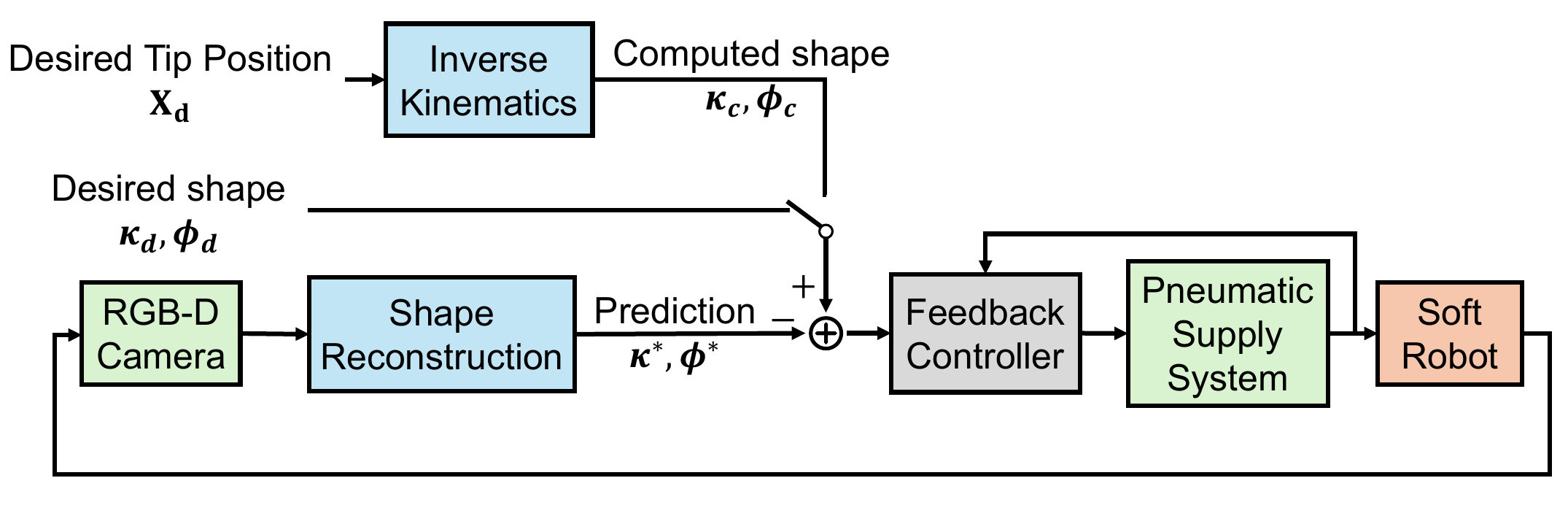}
    \caption{
    Closed-loop control framework. The system estimates the shape in real time and compares it with the desired configuration for feedback control.}
    \label{fig:controller}
\end{figure}

\begin{table}[ht]
\centering
\caption{Closed-loop control performance under different feedback sources.}
\label{tab:control_comparison}
\begin{tabular}{c c c c}
    \toprule
    \textbf{Task} & \textbf{Error Type} & \textbf{Feedback Source} & \textbf{Absolute Error (cm)} \\
    \midrule
    \multirow{2}{*}{Shape} 
        & \multirow{2}{*}{8-point avg.}
        & OptiTrack & $1.636 \pm 0.972$ \\
        &  
        & Ours    & $1.638 \pm 1.034$ \\
    \midrule
    \multirow{2}{*}{Tip} 
        & \multirow{2}{*}{End-point}
        & OptiTrack & $3.017 \pm 2.079$ \\
        &  
        & Ours    & $3.287 \pm 1.649$ \\
    \bottomrule
\end{tabular}
\end{table}

Representative closed-loop control results are shown in Fig.~\ref{fig:ctrl_results}. Table~\ref{tab:control_comparison} compares control performance using our reconstruction-based feedback versus OptiTrack baseline. The results show that our method achieves comparable accuracy, with low tracking errors in both shape and tip control tasks. Minor discrepancies are attributed to factors such as unmodeled nonlinear elongation at high pressure and deviations from the PCC assumption~\cite{mei2025learning}.

\begin{figure}[htbp]
    \centering
    \footnotesize
    \begin{subfigure}{0.45\columnwidth}
        \includegraphics[width=\linewidth]{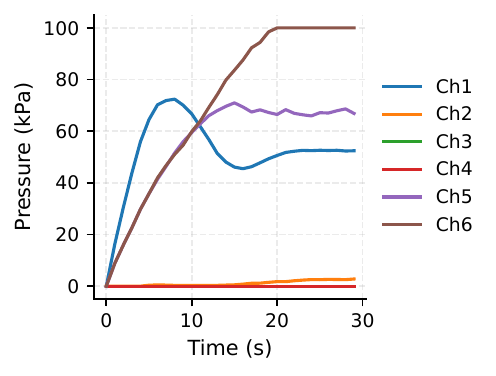}
        \caption{Input pressures.}
    \end{subfigure}
    \hfill
    \begin{subfigure}{0.45\columnwidth}
        \includegraphics[width=\linewidth]{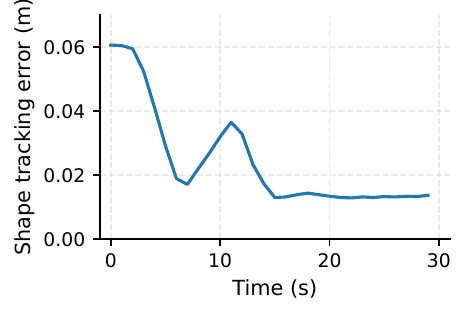}
        \caption{Shape tracking error.}
    \end{subfigure}
    \vskip 4pt
    \begin{subfigure}{0.45\columnwidth}
        \includegraphics[width=\linewidth]{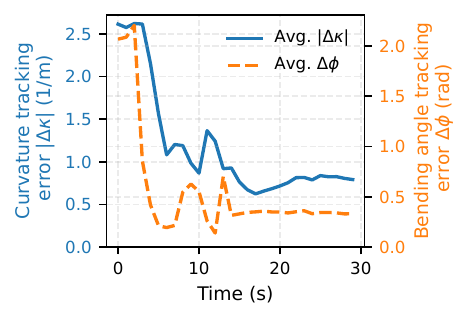}
        \caption{\(|\Delta\kappa|\) and \(\Delta\phi\) errors.}
    \end{subfigure}
    \hfill
    \begin{subfigure}{0.45\columnwidth}
        \includegraphics[width=\linewidth]{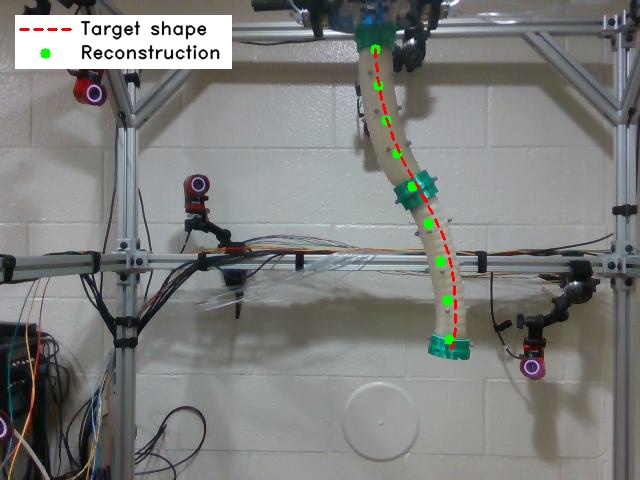}
        \caption{Visualization.}
    \end{subfigure}
    \caption{Closed-loop control demonstration. 
    (a) Six-chamber input pressures (Chamber 1-3: Segment~1, Chamber 4-6: Segment~2).
    (b) Shape tracking error over time.
    (c) Curvature $|\Delta\kappa|$ and bending-angle $\Delta\phi$ tracking errors (averaged over two segments).
    (d) Final snapshot: reconstructed shape vs target.}
    \label{fig:ctrl_results}
\end{figure}

\subsection{Demonstrations in Unstructured Environments}

We qualitatively evaluate our method in unstructured indoor and outdoor settings. As shown in Fig.~\ref{fig:outdoor_demo}, the system runs reliably across diverse conditions—including lighting variation, background clutter, and partial occlusion—without retraining or environment-specific tuning. These results highlight the method’s robustness and practicality for real-world deployment. Additional demonstration videos are provided in the supplementary material.

\section{CONCLUSIONS}
\label{sec:conclusion}

We presented a markerless, training-free method for real-time 3D shape reconstruction of soft continuum robots from RGB-D input. \texthlEdit{The method's core advantages lie in its minimal setup requirements and strong adaptability. It requires only an initial 3D reconstruction without the need for manual marker design, extensive training data, or environment-specific tuning. Moreover, the system enables robust shape reconstruction under partial occlusion, background clutter, and varying viewpoints.} These characteristics make the system highly suitable for rapid and low-cost deployment across diverse real-world environments. While our experiments focus on continuum robots, the method itself is configuration-agnostic and potentially applicable to a broader class of soft robots. However, the method assumes stable surface textures, potentially limiting performance on smooth or textureless bodies. Future work could explore integrating explicit markers or learning-based appearance models to enhance robustness.

\addtolength{\textheight}{-12cm}   




\bibliographystyle{IEEEtran}
\bibliography{refs}

@article{li2024unifying,
  title={Unifying {3D} representation and control of diverse robots with a single camera},
  author={Li, Sizhe Lester and Zhang, Annan and Chen, Boyuan and Matusik, Hanna and Liu, Chao and Rus, Daniela and Sitzmann, Vincent},
  journal={arXiv preprint arXiv:2407.08722},
  year={2024}
}

@article{albeladi2022hybrid,
  title={Hybrid eye-in-hand/eye-to-hand image based visual servoing for soft continuum arms},
  author={AlBeladi, Ali and Ripperger, Evan and Hutchinson, Seth and Krishnan, Girish},
  journal={IEEE Robotics and Automation Letters},
  volume={7},
  number={4},
  pages={11298--11305},
  year={2022},
  publisher={IEEE}
}

@article{kara2023towards,
  title={Towards design and development of an arUco markers-based quantitative surface tactile sensor},
  author={Kara, Ozdemir Can and Everson, Charles and Alambeigi, Farshid},
  journal={arXiv preprint arXiv:2310.08398},
  year={2023}
}

@article{zhang2019calibration,
  title={Calibration and external force sensing for soft robots using an {RGB-D} camera},
  author={Zhang, Zhongkai and Petit, Antoine and Dequidt, Jeremie and Duriez, Christian},
  journal={IEEE Robotics and Automation Letters},
  volume={4},
  number={3},
  pages={2356--2363},
  year={2019},
  publisher={IEEE}
}

@inproceedings{bern2020soft,
  title={Soft robot control with a learned differentiable model},
  author={Bern, James M and Schnider, Yannick and Banzet, Pol and Kumar, Nitish and Coros, Stelian},
  booktitle={2020 3rd IEEE International Conference on Soft Robotics (RoboSoft)},
  pages={417--423},
  year={2020},
  organization={IEEE}
}

@inproceedings{reiter2011learning,
  title={A Learning Algorithm for Visual Pose Estimation of Continuum Robots. In 2011 IEEE},
  author={Reiter, A and Goldman, RE and Bajo, A and Iliopoulos, K and Simaan, N and Allen, PK},
  booktitle={IEEE/RSJ International Conference on Intelligent Robots and Systems (IROS)},
  year={2011},
  pages={2390--2396}
}

@article{hoffmann2024iterative,
  title={An iterative closest point algorithm for marker-free {3D} shape registration of continuum robots},
  author={Hoffmann, Matthias K and M{\"u}hlenhoff, Julian and Ding, Zhaoheng and Sattel, Thomas and Fla{\ss}kamp, Kathrin},
  journal={arXiv preprint arXiv:2405.15336},
  year={2024}
}

@article{schindler2024image,
  title={Image-based backbone reconstruction for non-slender soft robots},
  author={Schindler, Leon and de Payrebrune, Kristin M},
  journal={PAMM},
  volume={24},
  number={4},
  pages={e202400130},
  year={2024},
  publisher={Wiley Online Library}
}

@inproceedings{lu2023image,
  title={Image-based pose estimation and shape reconstruction for robot manipulators and soft, continuum robots via differentiable rendering},
  author={Lu, Jingpei and Liu, Fei and Girerd, C{\'e}dric and Yip, Michael C},
  booktitle={2023 IEEE International Conference on Robotics and Automation (ICRA)},
  pages={560--567},
  year={2023},
  organization={IEEE}
}

@inproceedings{lal2021scoopnet,
  title={ScoopNet: 6DOF pose estimation pipeline for origami-inspired worm robots},
  author={Lal, Rohit and Swaminathan, Ruphan and Seenivasan, Lalithkumar and Qiu, Liang and Ren, Hongliang},
  booktitle={2021 IEEE International Conference on Development and Learning (ICDL)},
  pages={1--6},
  year={2021},
  organization={IEEE}
}

@article{rong2024vision,
  author={Rong, Yu and Gu, Guoying},
  journal={IEEE Robotics and Automation Letters}, 
  title={Vision-Based Real-Time Shape Estimation of Self-Occluding Soft Parallel Robots Using Neural Networks}, 
  year={2024},
  volume={9},
  number={8},
  pages={7349-7356},
  doi={10.1109/LRA.2024.3421794}}

@inproceedings{albeladi2021vision,
  title={Vision-based shape reconstruction of soft continuum arms using a geometric strain parametrization},
  author={AlBeladi, Ali and Krishnan, Girish and Belabbas, Mohamed-Ali and Hutchinson, Seth},
  booktitle={2021 IEEE International Conference on Robotics and Automation (ICRA)},
  pages={11753--11759},
  year={2021},
  organization={IEEE}
}

@article{zheng2024vision,
  title={Vision-based online key point estimation of deformable robots},
  author={Zheng, Hehui and Pinzello, Sebastian and Cangan, Barnabas Gavin and Buchner, Thomas JK and Katzschmann, Robert K},
  journal={Advanced Intelligent Systems},
  volume={6},
  number={10},
  pages={2400105},
  year={2024},
  publisher={Wiley Online Library}
}

@article{ravi2024sam,
  title={Sam 2: Segment anything in images and videos},
  author={Ravi, Nikhila and Gabeur, Valentin and Hu, Yuan-Ting and Hu, Ronghang and Ryali, Chaitanya and Ma, Tengyu and Khedr, Haitham and R{\"a}dle, Roman and Rolland, Chloe and Gustafson, Laura and others},
  journal={arXiv preprint arXiv:2408.00714},
  year={2024}
}

@inproceedings{deng2009imagenet,
  title={Imagenet: A large-scale hierarchical image database},
  author={Deng, Jia and Dong, Wei and Socher, Richard and Li, Li-Jia and Li, Kai and Fei-Fei, Li},
  booktitle={2009 IEEE Conference on Computer Vision and Pattern Recognition (CVPR)},
  pages={248--255},
  year={2009},
  organization={IEEE}
}

@inproceedings{he2016deep,
  title={Deep residual learning for image recognition},
  author={He, Kaiming and Zhang, Xiangyu and Ren, Shaoqing and Sun, Jian},
  booktitle={Proceedings of the IEEE conference on Computer Vision and Pattern Recognition (CVPR)},
  pages={770--778},
  year={2016}
}

@article{webster2010design,
  title={Design and kinematic modeling of constant curvature continuum robots: A review},
  author={Webster III, Robert J and Jones, Bryan A},
  journal={The International Journal of Robotics Research},
  volume={29},
  number={13},
  pages={1661--1683},
  year={2010},
  publisher={SAGE Publications Sage UK: London, England}
}

@inproceedings{schonberger2016structure,
  title={Structure-from-motion revisited},
  author={Schonberger, Johannes L and Frahm, Jan-Michael},
  booktitle={Proceedings of the IEEE Conference on Computer Vision and Pattern Recognition (CVPR)},
  pages={4104--4113},
  year={2016}
}

@article{kuhn1955hungarian,
  title={The Hungarian method for the assignment problem},
  author={Kuhn, Harold W},
  journal={Naval Research Logistics Quarterly},
  volume={2},
  number={1-2},
  pages={83--97},
  year={1955},
  publisher={Wiley Online Library}
}

@misc{gy920_segment_anything_2024,
  author       = {Gy920},
  title        = {Segment-anything-2-real-time},
  year         = {2024},
  note         = {Available: \url{https://github.com/Gy920/segment-anything-2-real-time}, Accessed: 2025-03-25}
}

@article{wade2022applications,
  title={Applications and limitations of current markerless motion capture methods for clinical gait biomechanics},
  author={Wade, Logan and Needham, Laurie and McGuigan, Polly and Bilzon, James},
  journal={PeerJ},
  volume={10},
  pages={e12995},
  year={2022},
  publisher={PeerJ Inc.}
}

@article{huang2021kinematic,
  title={Kinematic modeling and control of variable curvature soft continuum robots},
  author={Huang, Xinjia and Zou, Jiang and Gu, Guoying},
  journal={IEEE/ASME Transactions on Mechatronics},
  volume={26},
  number={6},
  pages={3175--3185},
  year={2021},
  publisher={IEEE}
}

@inproceedings{lin2017feature,
  title={Feature pyramid networks for object detection},
  author={Lin, Tsung-Yi and Doll{\'a}r, Piotr and Girshick, Ross and He, Kaiming and Hariharan, Bharath and Belongie, Serge},
  booktitle={Proceedings of the IEEE Conference on Computer Vision and Pattern Recognition (CVPR)},
  pages={2117--2125},
  year={2017}
}

@article{qi2017pointnet++,
  title={Pointnet++: Deep hierarchical feature learning on point sets in a metric space},
  author={Qi, Charles Ruizhongtai and Yi, Li and Su, Hao and Guibas, Leonidas J},
  journal={Advances in Neural Information Processing Systems (NeurIPS)},
  volume={30},
  year={2017}
}

@inproceedings{he2020momentum,
  title={Momentum contrast for unsupervised visual representation learning},
  author={He, Kaiming and Fan, Haoqi and Wu, Yuxin and Xie, Saining and Girshick, Ross},
  booktitle={Proceedings of the IEEE/CVF Conference on Computer Vision and Pattern Recognition (CVPR)},
  pages={9729--9738},
  year={2020}
}

@inproceedings{camarillo2008vision,
  title={Vision based 3-D shape sensing of flexible manipulators},
  author={Camarillo, David B and Loewke, Kevin E and Carlson, Christopher R and Salisbury, J Kenneth},
  booktitle={2008 IEEE International Conference on Robotics and Automation (ICRA)},
  pages={2940--2947},
  year={2008},
  organization={IEEE}
}

@article{vandini2017unified,
  title={Unified tracking and shape estimation for concentric tube robots},
  author={Vandini, Alessandro and Bergeles, Christos and Glocker, Ben and Giataganas, Petros and Yang, Guang-Zhong},
  journal={IEEE Transactions on Robotics},
  volume={33},
  number={4},
  pages={901--915},
  year={2017},
  publisher={IEEE}
}

@inproceedings{pedari2019spatial,
  title={Spatial shape estimation of a tendon-driven continuum robotic arm using a vision-based algorithm},
  author={Pedari, Yasaman and Parvaresh, Aida and Moosavian, S Ali A},
  booktitle={2019 7th International Conference on Robotics and Mechatronics (ICRoM)},
  pages={625--630},
  year={2019},
  organization={IEEE}
}

@article{elfferich2022soft,
  title={Soft robotic grippers for crop handling or harvesting: A review},
  author={Elfferich, Johannes F and Dodou, Dimitra and Della Santina, Cosimo},
  journal={IEEE Access},
  volume={10},
  pages={75428--75443},
  year={2022},
  publisher={IEEE}
}

@article{runciman2019soft,
  title={Soft robotics in minimally invasive surgery},
  author={Runciman, Mark and Darzi, Ara and Mylonas, George P},
  journal={Soft Robotics},
  volume={6},
  number={4},
  pages={423--443},
  year={2019},
  publisher={Mary Ann Liebert, Inc., publishers 140 Huguenot Street, 3rd Floor New~…}
}

@article{pan2022soft,
  title={Soft actuators and robotic devices for rehabilitation and assistance},
  author={Pan, Min and Yuan, Chenggang and Liang, Xianrong and Dong, Tianyun and Liu, Tao and Zhang, Junhui and Zou, Jun and Yang, Huayong and Bowen, Chris},
  journal={Advanced Intelligent Systems},
  volume={4},
  number={4},
  pages={2100140},
  year={2022},
  publisher={Wiley Online Library}
}

@article{della2023model,
  title={Model-based control of soft robots: A survey of the state of the art and open challenges},
  author={Della Santina, Cosimo and Duriez, Christian and Rus, Daniela},
  journal={IEEE Control Systems Magazine},
  volume={43},
  number={3},
  pages={30--65},
  year={2023},
  publisher={IEEE}
}

@article{mei2024simultaneous,
  author={Mei, Yu and Peng, Lei and Shi, Hongyang and Qi, Xinda and Deng, Yiming and Srivastava, Vaibhav and Tan, Xiaobo},
  journal={IEEE/ASME Transactions on Mechatronics}, 
  title={Simultaneous Shape Reconstruction and Force Estimation of Soft Bending Actuators Using Distributed Inductive Curvature Sensors}, 
  year={2024},
  volume={29},
  number={4},
  pages={2849-2857},
  doi={10.1109/TMECH.2024.3397825}}

@article{magistri2022contrastive,
  title={Contrastive {3D} shape completion and reconstruction for agricultural robots using {RGB-D} frames},
  author={Magistri, Federico and Marks, Elias and Nagulavancha, Sumanth and Vizzo, Ignacio and L{\"a}ebe, Thomas and Behley, Jens and Halstead, Michael and McCool, Chris and Stachniss, Cyrill},
  journal={IEEE Robotics and Automation Letters},
  volume={7},
  number={4},
  pages={10120--10127},
  year={2022},
  publisher={IEEE}
}

@article{mei2025learning,
  title={Learning-Based Modeling of Soft Actuators Using Euler Spiral-Inspired Curvature},
  author={Mei, Yu and Yuan, Shangyuan and Qi, Xinda and Fairchild, Preston and Tan, Xiaobo},
  journal={arXiv preprint arXiv:2504.18692},
  year={2025}
}

@article{habibian2022evaluation,
  title={Evaluation of two complementary modeling approaches for fiber-reinforced soft actuators},
  author={Habibian, Soheil and Wheatley, Benjamin B and Bae, Suehye and Shin, Joon and Buffinton, Keith W},
  journal={ROBOMECH Journal},
  volume={9},
  number={1},
  pages={12},
  year={2022},
  publisher={Springer}
}

\end{document}